\pdfoutput=1

\documentclass[11pt]{article}

\usepackage{ACL2023}

\usepackage{times}
\usepackage{latexsym}

\usepackage[T1]{fontenc}

\usepackage[utf8]{inputenc}

\usepackage{microtype}

\usepackage{inconsolata}

\usepackage{paralist}
\usepackage{adjustbox}

\usepackage{amsmath,amssymb,amsfonts}
\usepackage{pifont}
\usepackage{utfsym}

\usepackage{multirow}
\usepackage{booktabs}
\usepackage{arydshln}
\usepackage{bm}
\usepackage{color,xcolor}
\usepackage{colortbl}
\usepackage{soul}

\usepackage{tcolorbox}

\usepackage{adjustbox}

\newcommand{\mcolorbox}[2]{%
  \begingroup\setlength{\fboxsep}{1pt}%
  \colorbox{#1}{\text{\hspace*{2pt}\vphantom{Ay}#2\hspace*{2pt}}}%
  \endgroup
}

\newenvironment{myquote}%
  {\list{}{\leftmargin=0.3in\rightmargin=0.3in}\item[]}%
  {\endlist}

\definecolor{nmgray}{RGB}{229,229,229}
\definecolor{underlinegray}{RGB}{197,197,197}

\definecolor{introblue}{RGB}{0,176,240}
\definecolor{introgreen}{RGB}{0,203,134}
\definecolor{introgreen2}{RGB}{139,243,206}

\setul{0.3ex}{0.2ex}
\setulcolor{underlinegray}

\newtcolorbox{mybox}[2][]{
width=\columnwidth,
colback = nmgray!75!white, 
colframe = nmgray!75!white, 
boxsep=0pt,left=10pt,right=10pt,top=0pt,bottom=0pt,
fontupper=\linespread{0.9}\selectfont,
title=#2,#1}

\title{Reasoning Implicit Sentiment with Chain-of-Thought Prompting\Thanks{ The work is substantially supported by Alibaba Group through the Alibaba Innovative Research (AIR) Program.}
}

\author{Hao Fei$^1$, \, Bobo Li$^2$, \, Qian Liu$^3$,\,  Lidong Bing$^4$, \, Fei Li$^{2}$\Thanks{ Corresponding author: Fei Li.}, \, Tat-Seng Chua$^1$ \\
$^1$Sea-NExT Joint Lab, School of Computing, National University of Singapore \\
$^2$Key Laboratory of Aerospace Information Security and Trusted Computing,\\ 
Ministry of Education, School of Cyber Science and Engineering, Wuhan University\\ 
$^3$Sea AI Lab, \quad  $^4$DAMO Academy, Alibaba Group\\
\texttt{{haofei37@nus.edu.sg, boboli@whu.edu.cn, liuqian@sea.com,}}\\
\texttt{{l.bing@alibaba-inc.com, lifei\_csnlp@whu.edu.cn, dcscts@nus.edu.sg}}
}

\begin{document}
\maketitle
\begin{abstract}
While sentiment analysis systems try to determine the sentiment polarities of given targets based on the key opinion expressions in input texts, 
in implicit sentiment analysis (ISA) the opinion cues come in an implicit and obscure manner.
Thus detecting implicit sentiment requires the common-sense and multi-hop reasoning ability to infer the latent intent of opinion.
Inspired by the recent chain-of-thought (CoT) idea, in this work we introduce a \emph{Three-hop Reasoning} (\textsc{ThoR}) CoT framework to mimic the human-like reasoning process for ISA.
We design a three-step prompting principle for \textsc{ThoR} to step-by-step induce the implicit aspect, opinion, and finally the sentiment polarity.
Our \textsc{ThoR}+Flan-T5 (11B) pushes the state-of-the-art (SoTA) by over 6\% F1 on supervised setup.
More strikingly, \textsc{ThoR}+GPT3 (175B) boosts the SoTA by over 50\% F1 on zero-shot setting.
Our code is open at \url{https://github.com/scofield7419/THOR-ISA}.
\end{abstract}

\section{Introduction}

\vspace{-2mm}
Sentiment analysis (SA) aims to detect the sentiment polarity towards a given target based on the input text.
SA can be classified into explicit SA (ESA) and implicit SA (ISA), where the former type is the current mainstream task, in which the emotional expressions explicitly occur in texts \cite{pontiki-etal-2014-semeval}.
Different from ESA, ISA is much more challenging, because in ISA the inputs contain only factual descriptions with no explicit opinion expression directly given \cite{russo-etal-2015-semeval}.
For example, given a text `\emph{Try the tandoori salmon!}', having no salient cue word, almost all existing sentiment classifier\footnote{
We pre-experiment with total 20 existing SA models.
} predicts a neutral polarity towards `\emph{the tandoori salmon}'.
Human can easily determine the sentiment states accurately, because we always grasp the real intent or opinion behind the texts.
Thus, without truly understanding \emph{how} the sentiment is aroused, traditional SA methods are ineffective to ISA.

\begin{figure}[!t]
\centering
\includegraphics[width=1\columnwidth]{illus/intro4.pdf}
\caption{
Detecting the explicit and implicit sentiment polarities towards \textcolor{introblue}{targets}.
Explicit \mcolorbox{introgreen2}{opinion expression} helps direct inference, 
while detecting implicit sentiment requires common-sense and multi-hop reasoning.
}
\label{intro}
\vspace{-4mm}
\end{figure}

In fact, it is critical to first discover the hidden opinion contexts to achieve accurate ISA.
For the explicit case\#1 in Fig. \ref{intro}, it is effortless to capture the overall sentiment picture (e.g., `\emph{environment}' is the aspect, `\emph{great}' is the opinion), and thus can precisely infer the \emph{positive} polarity towards the given target \emph{hotel}.
Inspired by such fine-grained sentiment spirit \cite{xue-li-2018-aspect,zhang-etal-2021-aspect-sentiment,xu-etal-2020-position}, we consider mining the implicit aspect and opinion states.
For the implicit case\#2 in Fig. \ref{intro}, if a model can first infer the key sentiment components, e.g., the latent aspect `\emph{taste}', latent opinion `\emph{good and worth trying}', the inference of final polarity can be greatly eased.
To reach the goal, the capabilities of \textbf{common-sense reasoning} (i.e., infer what is `\emph{tandoori salmon}') and \textbf{multi-hop reasoning} (i.e., infer the aspect and then the opinion) are indispensable.

\begin{figure*}[!t]
\centering
\includegraphics[width=0.99\textwidth]{illus/framework4.pdf}
\caption{
An illustration of our \textsc{ThoR} framework for three-hop reasoning of implicit sentiment.
}
\label{framework}
\vspace{-4mm}
\end{figure*}

Fortunately, the recent great triumph of pre-trained large-scale language models (LLMs) offers a promising solution.
On the one hand, LLMs have been found to carry very rich world knowledge, showing extraordinary ability on common-sense understanding \cite{paranjape-etal-2021-prompting,liu-etal-2022-generated}.
On the other hand, the latest chain-of-thought (CoT) idea has revealed the great potential of LMs' multi-hop reasoning \cite{CoT-1,CoT-2,zhang2023automatic}, where an LLM with some prompts can do chain-style reasoning impressively.
Built on top of all these successes, in this work we implement a \underline{Th}ree-h\underline{o}p \underline{R}easoning CoT framework (namely \textsc{ThoR}) for ISA.
Based on an LLM, we design three prompts for three steps of reasoning, each of which respectively infers 1) the fine-grained aspect of the given target, 2) the underlying opinion towards the aspect, and 3) the final polarity.
With such easy-to-hard incremental reasoning, the hidden contexts of the overall sentiment picture can be elicited step by step to achieve an easier prediction of final polarity, which effectively alleviates the difficulties of the task prediction.

To ensure the correctness of each reasoning step, we consider a self-consistency mechanism for CoT inspired by \citet{CoT-voting}, 
which is to select the candidate answers (at each step) with high voting consistency of inferred aspect and opinion.
For supervised fine-tuning setup, we further propose a reasoning revising method.
We use the intermediate reasoning answers as model inputs to predict the final labels, where the supervision from gold labels will teach LLM to generate more correct reasoning.
On supervised fine-tuning setup, our Flan-T5 based \textsc{ThoR} improves the current best-performing baseline by more than 6\% in F1 score, and such margins are further magnified on zero-shot setup.
Most strikingly, our GPT3-based \textsc{ThoR} with 175B parameters boosts the baseline to a high-to 51.10\% increase of F1 score.

To sum up, this work contributes a multi-hop reasoning solution for implicit sentiment detection, which helps to achieve impressive improvement over the traditional non-reasoning methods.
To our knowledge, this is the first attempt to successfully extend the CoT idea to the sentiment analysis community.
Our method is simple yet effective, and can be broadly applied to other similar NLP problems without much effort.

\vspace{-2mm}
\section{Three-hop Reasoning Framework}

\vspace{-2mm}
The task of SA (either ESA or ISA) is defined as:
given a sentence $X$ with a target term $t\subset X$, a model determines the sentiment polarity $y$ towards $t$, i.e., \emph{positive}, \emph{neutral} or \emph{negative}.
We solve the task using an off-the-shelf LLM with prompt.
For the standard prompt-based method, we can construct the following prompt template as LLM's input:
\vspace{-3pt}
\begin{mybox}\texttt
\texttt{Given the sentence $X$, what is the sentiment polarity towards $t$?}
\end{mybox}
\vspace{-3pt}
\noindent The LLM should return the answer via: $\hat{y}$=$\text{argmax} p$($y|X,t$).

\vspace{-1mm}
\subsection{Chain-of-Thought Prompting}

\vspace{-1mm}
Now we consider the CoT-style prompt \cite{CoT-1,CoT-3} method for multi-step reasoning.
Instead of directly asking LLM the final result of $y$, in our \textsc{ThoR} (cf. Fig. \ref{framework}) we hope the LLM infer the latent aspect and opinion information before answering the finale $y$.
We here define the intermediate aspect term $a$ and latent opinion expression $o$.
We construct the three-hop prompts as follows.

\vspace{-1mm}
\paragraph{Step 1.}
We first ask LLM what aspect $a$ is mentioned with the following template:
\vspace{-3pt}
\begin{mybox}\texttt
\texttt{$C_1$[Given sentence $X$], which specific aspect of $t$ is possibly mentioned?}
\end{mybox}
\vspace{-3pt}
\noindent $C_1$ is the first-hop prompt context.
This step can be formulated as $A$=$\text{argmax} p$($a|X,t$), where $A$ is the output text which explicitly mentions the aspect $a$.

\vspace{-2mm}
\paragraph{Step 2.}
Now based on $X$, $t$ and $a$, we ask LLM to answer in detail what would be the underlying opinion $o$ towards the mentioned aspect $a$:
\vspace{-3pt}
\begin{mybox}\texttt
\texttt{$C_2$[$C_1$,$A$]. Based on the common sense, what is the implicit opinion towards the mentioned aspect of $t$, and why?}
\end{mybox}
\vspace{-3pt}
\noindent $C_2$ is the second-hop prompt context which concatenates $C_1$ and $A$.
This step can be written as $O$=$\text{argmax} p$($o|X,t,a$), where $O$ is the answer text containing the possible opinion expression $o$.

\vspace{-1mm}
\paragraph{Step 3.}
With the complete sentiment skeleton ($X$, $t$, $a$ and $o$) as context, we finally ask LLM to infer the final answer of polarity $t$:
\vspace{-3pt}
\begin{mybox}\texttt
\texttt{$C_3$[$C_2$,$O$]. Based on the opinion, what is the sentiment polarity towards $t$?}
\end{mybox}
\vspace{-3pt}
\noindent $C_3$ is the third-hop prompt context.
We note this step as $\hat{y}$=$\text{argmax} p$($y|X,t,a,o$).

\vspace{-1mm}
\subsection{Enhancing Reasoning via Self-consistency}

\vspace{-1mm}
We further leverage the self-consistency mechanism \cite{CoT-voting,CoT-voting2} to consolidate the reasoning correctness.
Specifically, for each of three reasoning steps, we set the LLM decoder to generate multiple answers, each of which will likely to give varied predictions of aspect $a$, opinion $o$ as well as the polarity $y$.
At each step, those answers with high voting consistency of inferred $a$, $o$ or $y$ are kept.
We select the one with highest confidence as the context in next step.

\vspace{-1mm}
\subsection{Reasoning Revising with Supervision}

\vspace{-1mm}
We can also fine-tune our \textsc{ThoR} when the on-demand training set is available, i.e., supervised fine-tuning setup.
We devise a reasoning revising method.
Technically, at each step we construct a prompt by concatenating 1) initial context, 2) this step's reasoning answer text and 3) final question, and feed it into LLM to predict the sentiment label instead of going to the next step reasoning.
For example, at end of step-1, we can assemble a prompt: [$C_1$,$A$, `\emph{what is the sentiment polarity towards $t$?}'].
In the supervision of gold labels, the LLM will be taught to generate more correct intermediate reasoning that is helpful to the final prediction.

\begin{table}[!t]
  \centering
\fontsize{9}{12.5}\selectfont
\setlength{\tabcolsep}{1.3mm}
\resizebox{0.98\columnwidth}{!}{
\begin{tabular}{lcccc}
\hline

\multicolumn{1}{c}{\multirow{2}{*}{\textbf{}}}& \multicolumn{2}{c}{\textbf{Restaurant}} & \multicolumn{2}{c}{\textbf{Laptop}} \\
\cmidrule(r){2-3}\cmidrule(r){4-5}
& All& 	ISA& 	All & 	ISA\\
 \hline
\multicolumn{5}{l}{$\bullet$ \textbf{\emph{State-of-the-art baselines}}} \\
BERT+SPC$^{\dagger}$ (110M) &	 77.16 &	 65.54  &	73.45 &	 69.54 \\
BERT+ADA$^{\dagger}$ (110M) &	80.05  &	65.92 &  	74.18  &	 70.11 \\
BERT+RGAT$^{\dagger}$ (110M) &	81.35  &	67.79  & 	74.07  &	 72.99 \\
BERT$_{\text{Asp}}$+CEPT$^{\dagger}$ (110M) &	82.07 &	67.79 &	78.38 &	75.86 \\
BERT+ISAIV$^{\dagger}$ (110M) &	81.40 &	69.66 &	77.25 &	78.29 \\
BERT$_{\text{Asp}}$+SCAPT$^{\dagger}$ (110M) &	83.79 &	72.28 &	79.15 &	77.59 \\

\hline

\multicolumn{5}{l}{$\bullet$ \textbf{\emph{Prompt-based methods}}}\\
BERT+Prompt (110M) &	81.34 &	70.12 &	78.58 &	75.24 \\
Flan-T5+Prompt (250M) &	81.50 &	70.91 &	79.02 &	76.40 \\
Flan-T5+Prompt (11B) &	84.72 &	75.10 &	82.44 &	78.91 \\
\hline

\multicolumn{5}{l}{$\bullet$ \textbf{\emph{CoT-based methods}}}\\
Flan-T5+\textsc{ThoR} (250M) &	82.98 &	71.70 &	79.75 &	67.63 \\
Flan-T5+\textsc{ThoR} (11B)&\bf 87.45 &\bf 79.73 &\bf 85.16 &\bf 82.43 \\
\quad w/o SelfConsistency &	86.03 &	77.68 &	84.39 &	80.27 \\
\quad w/o Reason-Revising &	86.88 &	78.42 &	84.83 &	81.69 \\

\hline
\end{tabular}%
}
\vspace{-1mm}
\caption{
F1 results on supervised fine-tuning setup.
Best results are marked in bold.
Scores by model with ${\dagger}$ are copied from \citet{li-etal-2021-learning-implicit}.
}
\vspace{-5mm}
  \label{main}%
\end{table}%

\vspace{-1mm}
\section{Experiments}

\vspace{-2mm}
\paragraph{Setups}
We experiment on the benchmark SemEval14 Laptop and Restaurant datasets \cite{pontiki-etal-2014-semeval}, where all the instances are split into explicit and implicit sentiment by \citet{li-etal-2021-learning-implicit}.
Since the encoder-style BERT cannot generate texts to support CoT, we use encoder-decoder style Flan-T5\footnote{
\url{https://huggingface.co/docs/transformers/model_doc/flan-t5}} as our backbone LLM.
We also test with GPT3 \cite{BrownMRSKDNSSAA20} and ChatGPT \cite{ouyang2022training}.
We used four versions of Flan-T5: 250M (base), 780M (large), 3B (xl) and 11B (xxl), and four versions of GPT3: 350M, 1.3B, 6.7B and 175B.
Note that GPT3 does not release the model parameters, and we use it in the prompting manner via the API\footnote{\url{https://beta.openai.com/docs/models/gpt-3}}.
This also means that we cannot perform supervised fine-tuning with GPT3.
We compare with the current best-performing baselines, including: 
BERT+SPC \cite{devlin-etal-2019-bert},
BERT+ADA \cite{rietzler-etal-2020-adapt},
BERT+RGAT \cite{wang-etal-2020-relational},
BERT$_{\text{Asp}}$+CEPT \cite{li-etal-2021-learning-implicit},
BERT+ISAIV \cite{wang-etal-2022-causal} and 
BERT$_{\text{Asp}}$+SCAPT \cite{li-etal-2021-learning-implicit}.
We consider both the supervised fine-tuning and zero-shot setups.
We adopt the F1 as the evaluation metric.
On the few-shot setup, we re-implement the baselines via their source codes.
Our experiments are conducted with 4 NVIDIA A100 GPUs.

\vspace{-2mm}
\paragraph{Results on Supervised Fine-tuning}
The comparisons are shown in Table \ref{main}.
It is interesting to see that the BERT with prompt learning under-performs the SoTA baseline BERT$_{\text{Asp}}$+SCAPT.
Even the Flan-T5-base (250M) with double-size parameters fails to beat the SoTA.
BERT$_{\text{Asp}}$+SCAPT is pre-trained on the large-scale sentiment aspect-aware annotation data, thus showing strong capability on SA.
But with our \textsc{ThoR} CoT prompting, Flan-T5-base clearly outperforms SoTA.
Further, when using the larger LLM, i.e., with 11B parameters, we can find the vanilla prompt-based Flan-T5 surpasses the best baseline.
More prominently, Flan-T5-11B with \textsc{ThoR} shows significant boosts for ISA, i.e., 7.45\%(=79.73-72.28) on Restaurant and 5.84\%(=82.43-77.59) on Laptop, with average improvement of 6.65\%(7.45+5.84)/2 F1.
Also the ablations of the self-consistency and reasoning revising mechanisms indicate their importances in our \textsc{ThoR} method.

\begin{table}[!t]
  \centering
\fontsize{9}{12.5}\selectfont
 \setlength{\tabcolsep}{1.mm}
\resizebox{1\columnwidth}{!}{
\begin{tabular}{lcccc}
\hline

\multicolumn{1}{c}{\multirow{2}{*}{\textbf{}}}& \multicolumn{2}{c}{\textbf{Restaurant}} & \multicolumn{2}{c}{\textbf{Laptop}} \\
\cmidrule(r){2-3}\cmidrule(r){4-5}
& All& 	ISA& 	All & 	ISA\\
 \hline
\multicolumn{5}{l}{$\bullet$ \textbf{\emph{State-of-the-art baselines}}} \\
BERT+SPC (110M) &	21.76 &	19.48 &	25.34 &	17.71 \\
BERT+RGAT (110M) &	27.48 &	22.04 &	25.68 &	18.26 \\
BERT$_{\text{Asp}}$+SCAPT (110M) &	30.02 &	25.49 &	25.77 &	13.70 \\
\hline

\multicolumn{5}{l}{$\bullet$ \textbf{\emph{Prompt-based methods}}}\\
BERT+Prompt (110M) &	33.62 &	31.46 &	35.17 &	22.86 \\
Flan-T5+Prompt (250M) &	54.38 &	41.57 &	52.06 &	31.43 \\
Flan-T5+Prompt (11B) &	57.12 &	45.31 &	54.14 &	33.71 \\
\hline

\multicolumn{5}{l}{$\bullet$ \textbf{\emph{CoT-based methods}}}\\
Flan-T5+\textsc{ThoR} (250M) &	55.86 &	41.84 &	52.52 &	32.40 \\
Flan-T5+\textsc{ThoR} (3B) &	57.33 &	42.61 &	56.36 &	38.16 \\
Flan-T5+\textsc{ThoR} (11B) &	{61.87} &	{52.76} &	{58.27} &	{40.75} \\
Flan-T5+ZeroCoT (11B) &	56.58 &	47.41 &	55.53 &	35.67 \\
GPT3+\textsc{ThoR} (175B) &	\bf 81.96 &	\bf 76.55 &	\bf 76.04 &	\bf 73.12 \\

 \hline
    \end{tabular}%
    }
    \caption{
Model results on Zero-shot setting.
We re-implement the state-of-the-art baselines for the zero-shot performance.
`ZeroCoT' means prompting LLM with the zero-shot CoT, `\emph{let's think step by step}' \cite{BrownMRSKDNSSAA20}.
    }
    \vspace{-4mm}
  \label{tab:Zero-shot}%
\end{table}%

\vspace{-2mm}
\paragraph{Results on Zero-shot Reasoning}

In Table \ref{tab:Zero-shot} we compare the zero-shot performances.
We can find that the improvement of both prompt-based and CoT-based methods over the current SoTA baseline increases dramatically.
But overall, the CoT-based methods with our \textsc{ThoR} show much more significant improvement on ISA.
For example, our Flan-T5-11B \textsc{ThoR} system gives over 30\% F1 average improvement over the best-performing baseline (BERT$_{\text{Asp}}$+SCAPT) on two datasets.
Most strikingly, when \textsc{ThoR} is equipped into super-large LLM, i.e., GPT3-175B, we can observe the impressive improvement, near to the level by Flan-T5-11B \textsc{ThoR} in supervised setting as in Table \ref{main}.
Specifically, it boosts the SoTA results by 51.94\%(=81.96-30.02) on Restaurant and 50.27\%(=76.04-25.77) on Laptop, with an average 51.10\%(51.94+50.27)/2 F1 leap.

\begin{figure}[!t]
\centering
\includegraphics[width=0.98\columnwidth]{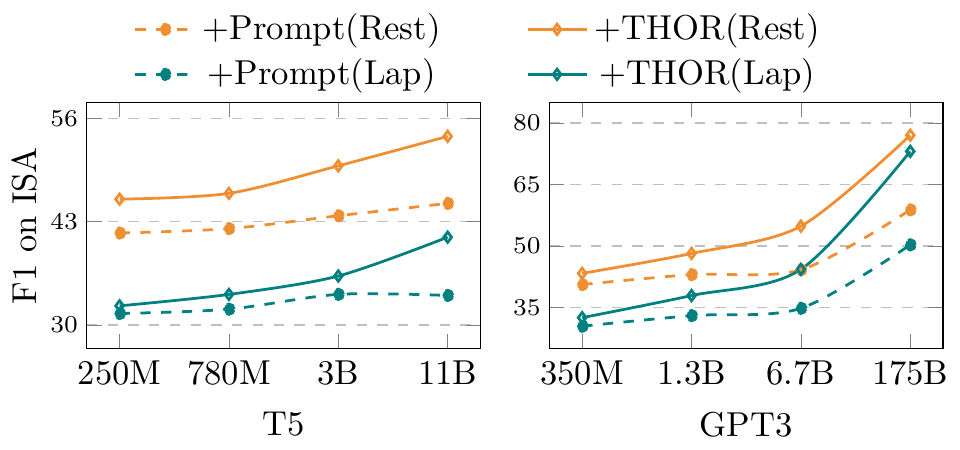}
\vspace{-2mm}
\caption{
Influences of LLM scales.
}
\vspace{-1mm}
\label{LM-scale}
\end{figure}

\begin{figure}[!t]
\centering
\includegraphics[width=0.98\columnwidth]{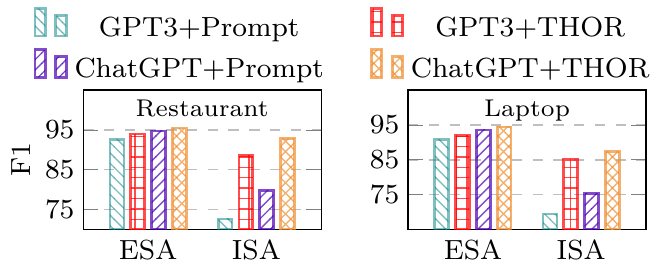}
\vspace{-2mm}
\caption{
Comparisons between GPT3\&ChatGPT on randomly-selected 50 ESA and 50 ISA instances.
}
\label{ChatGPT}
\vspace{-2mm}
\end{figure}

\vspace{-2mm}
\paragraph{Influence of Different Model Sizes of LLMs}
In Table \ref{main} and \ref{tab:Zero-shot} we have witnessed the power by using (very) large LLMs.
In Fig. \ref{LM-scale} we study the influence of different LLM scales.
We see that with the increasing model scale, the efficacy of our multi-hop reasoning prompting is exponentially amplified.
This coincides much with the existing findings of CoT prompting methods \cite{CoT-1,CoT-2,CoT-3}, i.e., the larger the LMs, the more significant improvement by CoT.
Because when the LLM is sufficiently large, the capabilities on common-sense and multi-hop reasoning are greatly developed and strengthened.

\vspace{-2mm}
\paragraph{Improving ChatGPT with \textsc{ThoR}}

The latest birth of ChatGPT has brought revolutionary advancement in NLP and AI community.
Here we compare the improvement of our \textsc{ThoR} on GPT3 (175B) and ChatGPT, respectively.
In Fig. \ref{ChatGPT} we show the testing results on 100 testing instances.
We can see that both LMs shows very high performances on ESA, and the enhancements by \textsc{ThoR} are very limited.
But prompting-based GPT3 and ChatGPT still fail much on ISA, where our \textsc{ThoR} has improved them on ISA very considerably.

\begin{figure}[!t]
\centering
\includegraphics[width=0.98\columnwidth]{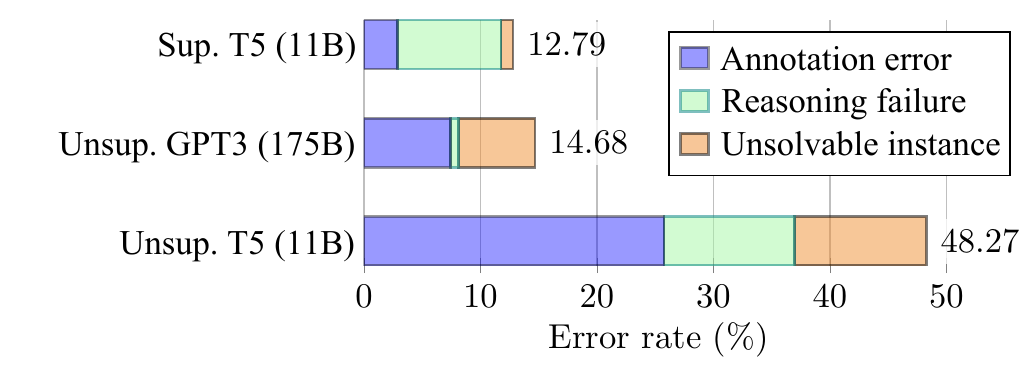}
\vspace{-2mm}
\caption{
Error analysis.
}
\label{error}
\end{figure}

\vspace{-1mm}
\paragraph{Failure Analysis}
In Fig. \ref{error} we show the error rates of failure cases when using \textsc{ThoR}, where we summarize three error types.
The Flan-T5-11B LLM gives 48.27\% error rate on zero-shot setup, while it goes down to 12.79\% when fine-tuned with supervision.
Unsupervised-GPT3 (175B) gives similarity low error rate as with Supervised-T5, while the latter fails much frequently on incapability of reasoning.
In contrast to Supervised-T5, the majority of failures in Unsupervised-GPT3 comes from the problematic data annotation.
Since Supervised-T5 is fine-tuned with supervision of `false' labels, it may actually learn the spurious correlations but with higher testing accuracy.

\vspace{-1mm}
\section{Related Work}

\vspace{-1mm}
Sentiment analysis has long been a hot research topic in NLP community \cite{PangL07,dong-etal-2014-adaptive,shi-etal-2022-effective}.
While the explicit SA models can make predictions based on the opinion expressions effortlessly, the implicit SA can be much more tricky due to the hidden opinion characteristics \cite{li-etal-2021-learning-implicit,wang-etal-2022-causal}.
And ISA is often more ubiquitous in realistic scenarios.
Although efforts have been made to ISA \cite{li-etal-2021-learning-implicit,wang-etal-2022-causal}, existing work can still be limited to the traditional paradigm of inference.
As aforementioned, ISA should be addressed via reasoning, i.e., common-sense and multi-hop reasoning.
Thus, this work follows such intuition, targeting solving ISA with a multi-hop reasoning mechanism.

As a key branch of SA, the fine-grained SA has been well explored \cite{WangPDX17,li-etal-2018-transformation,DiaASQ22bobo}.
The idea of fine-grained SA is to break down the SA into several key sentiment elements, including \emph{target}, \emph{aspect}, \emph{opinion} and \emph{sentiment polarity}, all of which together form a complete sentiment picture in detail \cite{PengXBHLS20,Feiijcai22UABSA}.
This work draws the same spirit of fine-grained SA.
We believe the reasoning of implicit sentiment should be an incremental process, inferring the sentiment elements step by step and finally understand the sentiment polarity in an easy-to-hard manner.

Language model pre-training has received increasing research attention for enhancing the utility of downstream applications \cite{RaffelSRLNMZLL20}
Most recently, the large-scale language models (LLMs) have shown great potential to the human-level intelligence, e.g., ChatGPT \cite{ouyang2022training}.
LLMs have extensively demonstrated to exhibit extraordinary abilities on common-sense understanding \cite{paranjape-etal-2021-prompting,liu-etal-2022-generated} and multi-hop reasoning \cite{CoT-1,CoT-2}.
This work implements the implicit sentiment reasoning built upon LMs, based on the latest proposed chain-of-thought (CoT) idea.
CoT prompting is a gradient-free technique that induces large LMs to produce intermediate reasoning steps leading to the final answer.
\citet{CoT-1} formally study the CoT prompting in language models, in which they elicit LMs to generate coherent series of intermediate reasoning steps that direct to the final answer to the original question.

\vspace{-1mm}
\section{Conclusion}

\vspace{-1mm}
In this paper, we present a \emph{Three-hop Reasoning} prompting framework to achieve the chain-of-thought reasoning process for implicit sentiment analysis.
Based on the existing LLM, we design three prompts for three steps of reasoning, each of which respectively infers the fine-grained aspect, the underlying opinion and the final polarity.
On the ISA datasets, different LLMs equipped with our \textsc{ThoR} show impressive performances over the existing best-performing baselines on both the supervised and zero-shot setups.
We show that the larger the LLMs, the more significant improvement by our \textsc{ThoR} method.


\section*{Acknowledgments}

The work is also partially supported by the National Key Research and Development Program of China (No. 2022YFB3103602) and the
Sea-NExT Joint Lab at National University of Singapore.



\section*{Limitations}

\textsc{ThoR} helps unleash the full power of LLMs only when being integrated into the large enough models, while on the middle or lower size LLMs, the improvement by \textsc{ThoR} will be limited to certain extent, due to the emergence nature of LLMs.

\bibliography{anthology}
\bibliographystyle{acl_natbib}

\newpage

\appendix


\section{Appendix}

Here we present several pieces of real testing examples.
We compare \textsc{ThoR} with the vanilla prompting method, and the zero-shot CoT method (Prompt + `\emph{Lets think step by step}').
We perform the comparisons based on the ChatGPT.\footnote{\url{https://chat.openai.com/}, Dec. 15, 2022}

\begin{figure*}[!t]
\centering
\includegraphics[width=0.9\textwidth]{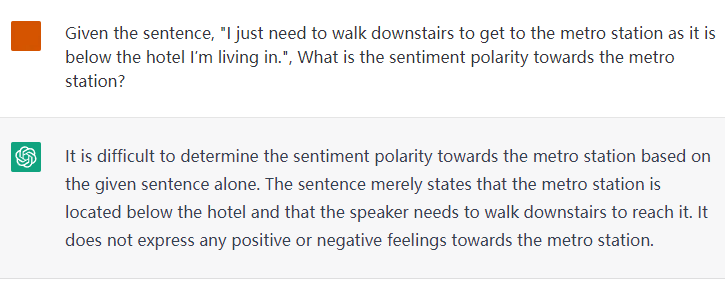}
\caption{
Vanilla prompt-based result for testing case-I.
}
\label{case1-prompt}
\end{figure*}

\begin{figure*}[!t]
\centering
\includegraphics[width=0.91\textwidth]{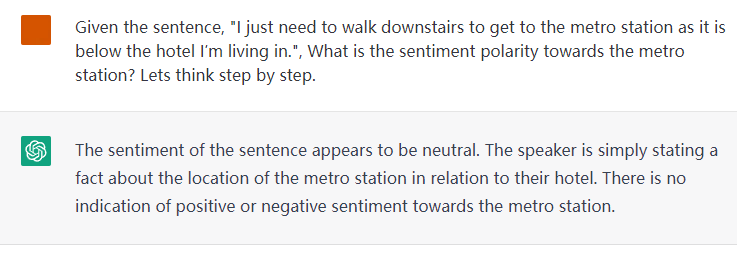}
\caption{
Result by zero-shot CoT method for testing case-I.
}
\label{case1-ZeroCoT}
\end{figure*}

\newpage

\paragraph{$\bullet$ Case-I}

Input text:
\begin{myquote}
\emph{I just need to walk downstairs to get to \textcolor{introblue}{the metro station} as it is below the hotel I’m living in.}
\end{myquote}
The gold sentiment label is \emph{positive} towards \emph{\textcolor{introblue}{the metro station}}.

\begin{figure*}[!t]
\centering
\includegraphics[width=0.91\textwidth]{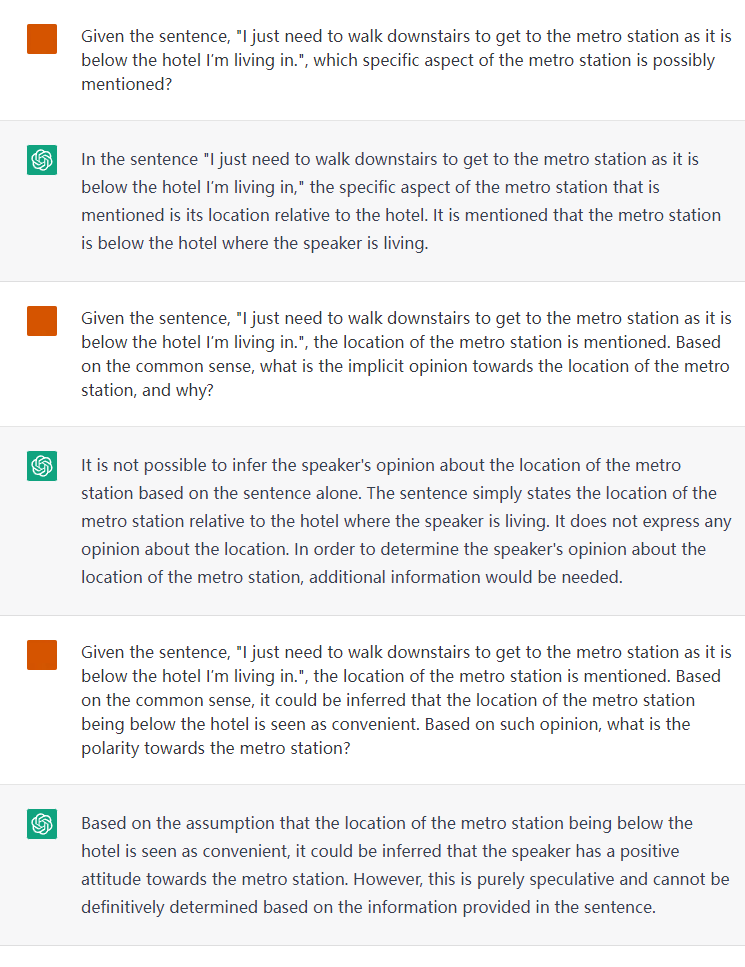}
\caption{
Result by our \textsc{ThoR} method for testing case-I.
}
\label{case1-THOR}
\end{figure*}

In Fig. \ref{case1-prompt}, \ref{case1-ZeroCoT} and \ref{case1-THOR}, we show that our \textsc{ThoR} successfully induces the ChatGPT to finally give a correct decision on sentiment polarity, where the other two methods fail.

\newpage

\paragraph{$\bullet$ Case-II}

Input text:
\begin{myquote}
\emph{\textcolor{introblue}{Lunch} came with pickels and slaw, no extra charge.}
\end{myquote}
The gold sentiment label is \emph{positive} towards \emph{\textcolor{introblue}{Lunch}}.

\begin{figure*}[!t]
\centering
\includegraphics[width=0.9\textwidth]{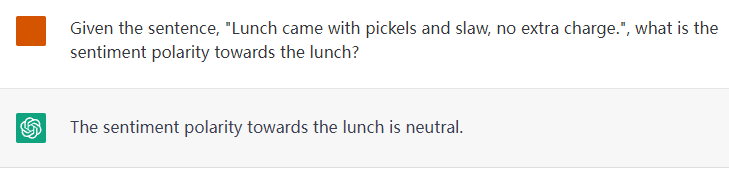}
\caption{
Vanilla prompt-based result for testing case-II.
}
\label{case2-prompt}
\end{figure*}

\begin{figure*}[!t]
\centering
\includegraphics[width=0.91\textwidth]{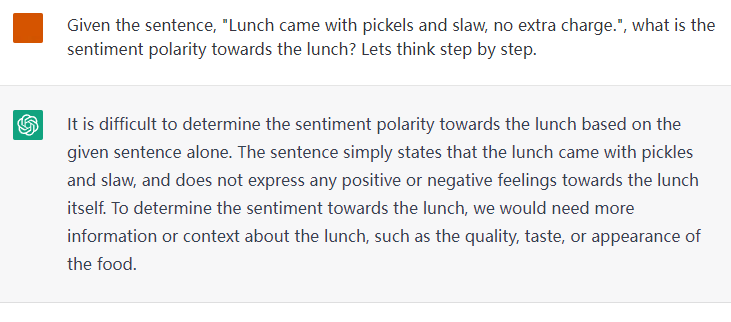}
\caption{
Result by zero-shot CoT method for testing case-II.
}
\label{case2-ZeroCoT}
\end{figure*}

\begin{figure*}[!t]
\centering
\includegraphics[width=0.91\textwidth]{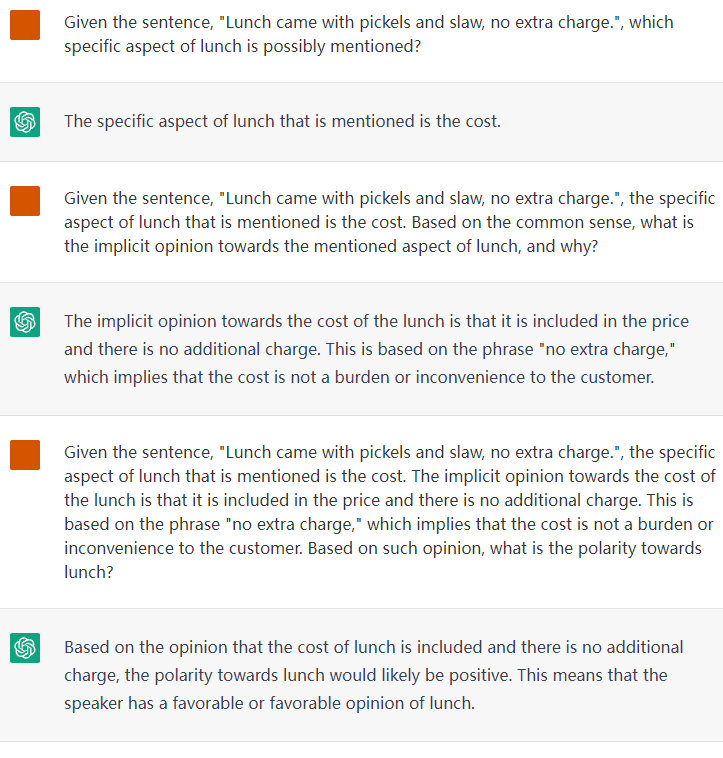}
\caption{
Result by our \textsc{ThoR} method for testing case-II.
}
\label{case2-THOR}
\end{figure*}

Fig. \ref{case2-prompt}, \ref{case2-ZeroCoT} and \ref{case2-THOR} shows the results and the LLM's response, respectively.
Our \textsc{ThoR} induces the ChatGPT to draw a correct decision on sentiment polarity, but the other two methods still fail.

\end{document}